\documentclass[letterpaper, 10 pt, conference]{ieeeconf}  % Comment this line out if you need a4paper

\IEEEoverridecommandlockouts                              % This command is only needed if 
\usepackage{graphicx} 
\usepackage{caption}
\usepackage{subcaption}
\usepackage{times} 
\usepackage{amsmath}
\usepackage{amssymb} 
\usepackage[english]{babel}
\usepackage{algorithm2e}
\usepackage{balance}
\usepackage{url}
\usepackage{xcolor}

\newcommand\reviewFernando[1]{{\color{black}#1}}
\newcommand\reviewLuis[1]{{\color{black}#1}}

\title{\LARGE \bf
DLL: Direct LIDAR Localization. A map-based localization approach for aerial robots
}

\author{Fernando Caballero$^{1}$ and Luis Merino$^{2}$% <-this % stops a space
\thanks{*This work was partially supported by the  Spanish Ministry of Science, Innovation and Universities (COMCISE RTI2018-100847-B-C22, MCIU/AEI/FEDER, UE) and by the Junta de Andalucía (DeepBot, PY20\_00817)}% <-this % stops a space
\thanks{$^{1}$Fernando Caballero is with the Service Robotics Laboratory, Universidad de Sevilla, Seville, Spain. {\tt\small fcaballero@us.es}}
\thanks{$^{2}$Luis Merino is with the Service Robotics Laboratory, Universidad Pablo de Olavide, Seville, Spain. {\tt\small lmercab@upo.es}}%
}

\begin{document}

\maketitle
\thispagestyle{empty}
\pagestyle{empty}

%%%%%%%%%%%%%%%%%%%%%%%%%%%%%%%%%%%%%%%%%%%%%%%%%%%%%%%%%%%%%%%%%%%%%%%%%%%%%%%%
\begin{abstract}
This paper presents DLL, a \reviewFernando{fast} direct map-based localization technique using 3D LIDAR for its application to aerial robots. DLL implements a point cloud to map registration based on non-linear optimization of the distance of the points and the map, thus not requiring  features, neither point correspondences. Given an initial pose, the method is able to track the pose of the robot by refining the predicted pose from odometry. Through benchmarks using real datasets and simulations, we show how the method performs much better than Monte-Carlo localization methods and achieves comparable precision to other optimization-based approaches but running one order of magnitude faster. The method is also robust under odometric errors. The approach has been implemented under the Robot Operating System (ROS), and it is publicly available. 
\end{abstract}

%%%%%%%%%%%%%%%%%%%%%%%%%%%%%%%%%%%%%%%%%%%%%%%%%%%%%%%%%%%%%%%%%%%%%%%%%%%%%%%%
\section{Introduction}

%Operations oriented to localization, that do not require SLAM or exploration. REFs 
%Specifically, indoor or GNSS-denied envionemnts
%Many robot applications require routinely visiting points in the environment to perform operations like periodic inspection, inventory in warehouses, load transportation and many others. In many of these applications it is possible to build a priori a map of the environment in which the robot will operate by using mapping and/or SLAM modules. In the absence of a global localization system, like GPS or artificial positioning devices as laser reflectors, this map can used for robot localization. If the map does not change often, after the initial map is built, pure localization typically suffices for robot operation. 

%Then UAVs, REFs 
%The quest for the equivalent to AMCL in 3D
%With the advent of Unmanned Aerial Vehicles (UAVs), the former operations are being extended to the third dimension. As such, there are UAV systems for warehousing applications \cite{kwon2019robust}, inspection  , etc. Like with ground robots in indoor environments, in which 2D map-based localization is at the heart of the operation of the robot (typically using Monte-Carlo Localization -MCL- methods), we need equivalent efficient 3D map-based localization systems to be used by UAVs. They should be fast and reliable, so that the localization information can be used by a robot to plan actions.

\begin{figure}[!t]
     \centering
         \includegraphics[width=0.8\columnwidth]{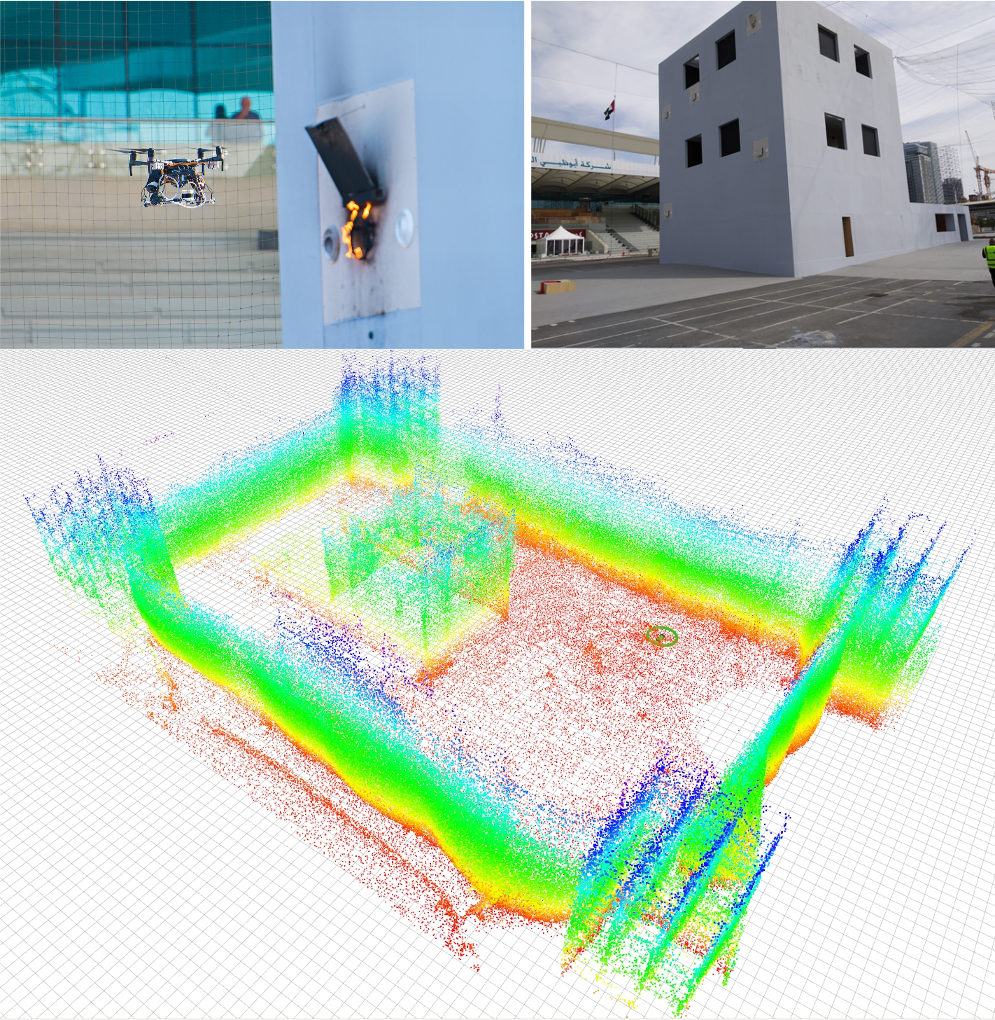}
         \caption{Scenario of the MBZIRC'20 competition, Challenge 3. UAVs have to put out fires (top, left) in a high-rise building scenario (top, right). A 3D map (bottom) built from 3D LIDAR data is used for localization.}
    \label{fig:mbzirc-arena}
\end{figure}

The operation of Unmanned Aerial Vehicles (UAVs) when there is no global localization system, like GPS, or other external positioning devices, like laser reflectors, calls for alternatives for estimating the robot pose. In applications where it is possible to build a priori a 3D map of the environment in which the UAV will operate, map-based localization can suffice if the environment does not change very often. This map can be created and updated by using SLAM modules (see Fig. \ref{fig:mbzirc-arena}). We need, thus, efficient 3D map-based localization systems to be used by UAVs. They should be fast and reliable, so that the localization information can be used by a robot to plan its actions.

%LIDAR-based. Why. Include refs
%While visual sensors, including RGBD, have been prominent for 3D perception in the UAV literature, mainly for their small weight and small power consumption, their limited field of view and range impose limitations. 
3D Light Detection and Ranging (LIDAR) systems are becoming smaller, cheaper and lighter, allowing UAVs to carry such sensors. They provide rich and precise long range 3D information that can be leveraged for localization. This paper presents a map-based position tracking  %localization system 
using 3D LIDAR measurements to be used in aerial robots. 

Among the different challenges involved, the 3D LIDAR registration (alignment with respect to the map) stands as the major technological problem to solve. Most of the approaches in the state of the art make use of geometric features such as corners or planes extracted from the sensor point cloud \cite{rs11111348,LOAM,ulas13}. The key idea is to select points or areas into the point cloud that are easy to identify and match when we see again the scene from different points of views and/or distances. These approaches are fast to compute, and accurate. However, feature-based 3D LIDAR-to-map registration is a difficult task \reviewFernando{due to different view-point perception and occlusions}. %Normally, the 3D map has much higher density of points than the LIDAR sensor readings. 
This forces to use approaches for feature detection \reviewFernando{and description in point-clouds with different densities \cite{FCGF2019} \cite{gojcic20193DSmoothNet}}, or to built virtual views of the map depending on the robot relative position. Also, feature-based approaches can struggle in scenarios with no distinctive regions.

Iterative closest points (ICP) approaches \cite{ChenM92} use the raw 3D point cloud of the sensor, and no features or interests points are extracted. Although they are very accurate, their convergence depends on the goodness of the initialization and, most importantly, they require lots of computation to deal with 3D large point clouds. The nearest point search is the main computational bottleneck of ICP approaches.

The Normal Distribution Transform (NDT) was one the first approaches to propose a nearest-neighbor-free solution for 3D registration \cite{magnusson2007scan,8206170}, reducing the computational time with respect to ICP approaches, while keeping the accuracy. The NDT is used to encode the scan and the map, using a probabilistic representation. Registration is then performed between these representations and not the original scans. This representation allows using numerical optimization methods, and does not require a nearest-neighbor search. Unfortunately, the numerical solution computation makes the approach difficult to compute in real-time when large point clouds are considered.

%The Normal Distribution Transform (NDT) has been used for 3D registration \cite{magnusson2007scan,8206170}. 

%The presented approach follows the NDT key idea of removing the nearest-neighbor search. Distance Function representation

Here we propose a direct approach for registration, and integrate it into a localization system called Direct LIDAR Localization (DLL). By direct we mean that the raw point cloud is used and no nearest point search is required. DLL estimates the transform that aligns the input point cloud and the map by minimizing the distances of points to the map using non-linear optimization. While this optimization would be computationally demanding, DLL takes advantage of the localization setup (in which you need an a priori map of the environment) to pre-compute a Distance Field (DF) \cite{jones20063d} map that boosts the optimization process. We show in this paper, using datasets from simulated and real flights, how this approach is more efficient and robust than alternative methods from the state of the art, achieving similar accuracy. The resultant system \reviewFernando{is about one order to magnitude faster than other approaches in the state of the art} and can be used for real-time position tracking of aerial robots using 3D LIDAR as main sensor. 

%The main contribution of the paper is then a novel, that

%These contributions are validated using datasets of UAV with onboard 3D LIDAR. The evaluation 

\section{Related Work}

%We actually do position tracking, so we have to be careful when mentioning global localization approaches

%Review
%The majority of approaches for 3D LIDAR-based localization are related to intelligent ground vehicles. 

A majority of works from the literature on 3D LIDAR-based localization deal with odometry and SLAM approaches, while there are less approaches dealing with map-based localization using 3D LIDARs. A recent review on localization approaches using 3D LIDAR for intelligent ground vehicles can be found in \cite{elhousni2020survey}. 

%Most odometry works

The basic procedure is to register the current LIDAR reading into the map (this map being the last LIDAR reading or a partial submap in the case of odometry).
%3D feature-based, also brief. But LOAM is there. Include some more
As mentioned before, many approaches first extract features from the point clouds and from the map that are then used for registration. One such approach is LOAM \cite{LOAM,zhang2017low-drift}. It uses point features on sharp edges and planar surfaces extracted from the 3D scans. These features are matched between consecutive scans for fast odometry, and between the scan and a map that is being built at a lower rate (without loop closures), leading to an enhanced odometry estimate. 
%Some more details and drawbacks
Another example is \cite{dube2018incremental}, where the 3D point clouds are first segmented, and the resulting segments are then used to find matches with respect to a map using a set of descriptors for localization. \reviewFernando{MULLS \cite{pan2021mulls} is another example of a feature-based LIDAR SLAM approach, which increases the accuracy of the estimation by combining a feature-based front-end SLAM with a multi-metric ICP for map alignment.} 

%However, feature-based 3D LIDAR to map registration is a difficult task. Normally, the 3D map has much higher density of points than the LIDAR sensor readings. This forces to use different approaches for feature detection in LIDAR and map, or to built virtual views of the map depending on the robot relative position.   

%Deep learning methods, brief. Drawbacks
Lately, learning-based methods are being also considered for the problem. 
%LocNet does global loc and also position tracking. Drawback?
LocNet \cite{yin2018locnet} defines a handcrafted representation of LIDAR frames and then learns a Siamese network architecture to estimate the similarity between the representations of local scans and scans stored in the map. The system is integrated into a Monte-Carlo Localization for 2D pose estimation, which is then refined to 3D pose estimation through ICP. In \cite{dube2020segmap}, the segments in \cite{dube2018incremental} are augmented with an efficient 3D segment descriptor learned from data for global mapping and localization. \reviewFernando{Learning has been also used for key-point detection and description \cite{Bai2020D3FeatJL}\cite{FCGF2019}\cite{gojcic20193DSmoothNet} in order to increase the robustness of feature-based LIDAR SLAM approaches and to generalize to different 3D point cloud densities.}

%Registration-based methods. ICP-based approaches, NDT
Closer to our approach are registration-based methods. These methods employ ICP and its variants directly over the point clouds, and are robust and accurate. However they use to be computationally expensive, preventing their use for real time systems. Approximate neighbors search has been proposed to significantly reduce computation in \cite{muja2009fast,elsebergcomparison}, and combined with a clever point selection a SLAM system can be built. However, their computation grows (but slowly) with the size of the map and it is not clear how to extend this configuration to map-based localization. 
In \cite{kovalenko2019sensor}, nearest point correspondences between consecutive scans are used to estimate 6 DOF odometry. The data association step here is crucial, as wrong correspondences can lead to erroneous pose estimations. The paper proposes point filters and geometric match rejection steps to reduce the computational burden and improve the quality of the registration process. \reviewFernando{Other ICP variants focus on improving the global robustness of the approach against noise or bad initialization by means of new criteria as Branch-and-Bound in GO-ICP \cite{YangPAMI2016}, or reformulating the ICP problem as Expectation-Maximization in EM-ICP \cite{GrangerECCV2002} or as a Truncated Lest Squares in TEASER \cite{YangTRO2021}. However, they can be computationally expensive.}

%Check if this is true
%Map-based localization-based on registration

%NDT-based localization. Important as several things are common
Direct methods that avoid the point association step present advantages in this regard. In that direction, the NDT representation discussed above has been used for map-based Monte-Carlo localization using LiDAR \cite{6696380}. The NDT representation is itself a likelihood model for the alignment of the cloud and the map. In \cite{sun2020localising}, Monte-Carlo localization using NDT is mixed with a deep probabilistic model learned from data that provides prior estimations for the filter. 

%Former work, AMCL3D
\reviewFernando{Distance Fields (DF) and its variants (Signed Distance Fields (SDF), Truncated Signed Fields (TSDF), Euclidean Signed Distance Fields (ESDF), etc) have emerged as a very convenient tool to represent the robot environment for mapping and planning applications \cite{Oleynikova2016SignedDF}. Thus, a pose tracking method based on 3D Monte Carlo localization and 3D DF for map representation is presented in \cite{Perezgrau17JARS}}. A particle filter is used to track the pose of the UAV. The likelihood function for each scan is derived from the DF, and no point correspondences are needed. A similar approach is considered in \cite{Akai2020}, where an efficient 3D DF representation is presented to compute those likelihoods. These approaches are robust, but significantly less precise than Gaussian filtering or optimization-based methods. %Also, the method considers potential unmapped measurements to improve the localization.
\reviewFernando{TSDF is also used to efficiently represent 3D submaps in a decentralized SLAM approach, presented in \cite{DuboisIros2020}. The submaps are shared among robots to find matches between them and refine inter-robot localization. However, the matching process needs to be improved in order to properly scale to a larger number of agents.}

\reviewFernando{Direct registration has been also proposed for 2D robot localization in \cite{Pedrosa2017EfficientLB}. This approach pre-computes a likelihood function over the closest distance to obstacles in a 2D map (as in \cite{HornungIROS2010} and \cite{Perezgrau17JARS} for 3D distance computation), and makes use of non-linear optimization to estimate the best robot position into the reference map. However, it is limited to 2D localization, extending to 3D requires significant improvements in order to minimize the  computation. Additionally, it optimizes a Gaussian function over the distances, which might slightly slow the optimization process.}

%Relate to it. Why not filtering and optimization
Here, we present a novel direct approach based on non-linear optimization instead of filtering. This has been shown to present advantages with respect to filter-based approaches for localization in 2D  \cite{dantanarayana2016c}. \reviewFernando{Our method} uses a 3D distance field and an efficient optimization  formulation to correct the predicted position given by odometry using the map. The resultant localization system only requires a LIDAR and IMU to operate.

\section{DLL: Direct LIDAR Localization}
\label{sec:method}

The process of map-based robot localization using point clouds can be summarized as finding the transform that better aligns the current LIDAR point cloud with respect to the map. However, assuming we are localizing a robot, DLL makes use of the following prior information to build a fast and accurate localization algorithm:

\begin{itemize}
    \item Robot odometry. This gives us a good prior about how much the robot moved since we localized the robot for the last time. This prior can be used to give a good guess of the robot position/orientation into the map, saving computational time.
    \item Inertial Measurement Units (IMUs). Nowadays, IMUs are cheap and accurate, they are installed in all smartphones. It is very common to have an IMU onboard the robot. Moreover, precise IMUs are mandatory for aerial robots in order to close the control loops. While yaw angle is subject to significant distortion due to magnetic field interference in the compass, roll and pitch angles are observable and accurate. This information can be used to tilt-compensate the LIDAR point cloud, simplifying the registration process.
\end{itemize}

%The integration of these sensor together with point-cloud registration with respect to a map can be implemented easily using approaches into the state of the art, as ICP [REF] or NDT [REF] for registration. However, these algorithms are computationally demanding, preventing their use in real time applications in 3D. 

DLL proposes a new point cloud registration process that is fast and accurate, so that in can be used as primary localization system in a real robot installation. Next paragraphs formulate the problem and provide detailed information about such registration process.

\subsection{Direct map to point cloud registration}

We assume the LIDAR (or any other 3D sensor) provides a point cloud \mbox{$\mathbf{pc}^k:\{\mathbf{p}_0,\mathbf{p}_1, ... \mathbf{p}_P\}$} at every time step $k$, where $\mathbf{p}_i=[px_i,py_i,pz_i]^t$. 

A map of the environment is also available as a point cloud \mbox{$\mathbf{map}:\{\mathbf{m}_0,\mathbf{m}_1, ... \mathbf{m}_M\}$}, where $\mathbf{m}_i=[mx_i,my_i,mz_i]^t$. This map is assumed to be static. 

Given the 3D map and the point cloud at time $k$, our objective is to compute the transform $\mathbf{T}_{map}^k$ that better aligns $\mathbf{pc}^k$ to the $\mathbf{map}$. This can be achieved by minimizing the following expression:

\begin{equation}
\label{eq:geneal_opt}
\operatorname*{arg\,min}_{\{\mathbf{T}_{map}\}}\left[ \sum_{i=1}^P\|\mathbf{T}_{map}\mathbf{p}_i-\mathbf{m}_c(\mathbf{p}_i)\|^{2}\right]
\end{equation}

\noindent where $\mathbf{m}_c(\mathbf{p}_i)$ is the point of $\mathbf{map}$ closest to $\mathbf{p}_i$.

Solving (\ref{eq:geneal_opt}) needs to deal with two major challenges: how to determine the values of $\mathbf{m}_c$, and how to solve a massively overdetermined non-linear optimization problem.

ICP works by iteratively searching for pairs of nearby points in the two point clouds and minimising the sum of all point-to-point distances. The main bottleneck is the nearest neighbor computation. Approximately nearest search or kd-trees can be used to speed up the process, but still requires significant computational resources.

On the other hand, NDT models the point cloud as a combination of normal distributions instead of individual points, describing the probability of finding a point at a certain position. This is a piecewise smooth representation of the point cloud, which enables standard numerical optimization methods for registration. This avoids explicit neighbor search, with the corresponding computational reduction. 

DLL follows NDT's key idea of modeling the registration process as a non-linear optimization process, but models the point cloud as a distance field. Instead of individual points, DLL builds a distance field representation of the map called $DF(\mathbf{x})$. Given a 3D point $\mathbf{x}=[x, y, z]^t$, $DF(\mathbf{x})$ provides the distance to the closest point into the map. This way, we transform (\ref{eq:geneal_opt}) into the following expression:

\begin{equation}
\label{eq:dll_opt}
\operatorname*{arg\,min}_{\{\mathbf{T}_{map}\}}\left[ \sum_{i=1}^P DF^{2}(\mathbf{T}_{map}\mathbf{p}_i)\right]
\end{equation}

\noindent $DF(\mathbf{x})$ is continuous and generally smooth (except in the cut locus and object boundaries, where the gradient is discontinuous \cite{jones20063d}), so we can use regular optimization processes to solve the expression. 

Estimating the corresponding \reviewFernando{distance field} of the 3D map can be computationally demanding, specially for large maps. However, this only needs to be computed once, and we can compute it offline as part of the map building process for map-based localization. 

Next Section details the process of approximating the distance function of the 3D point cloud used as map.

\subsection{Approximated distance field}

The distance function is computed from the 3D map and stored as a 3D grid of fixed resolution. The size of the grid is determined by the spacial distribution of 3D points. Every grid cell contains the distance to the closest point of the 3D map to that cell.

The 3D grid is build by systematically computing the distance to the closest map point in all the cells. We use a kd-tree in order to speed up the computation, although other methods for approximately nearest point search can be applied. This process is computationally expensive, but it is performed offline prior to robot localization.

Once built, trilinear interpolation parameters are computed for each cell. The eight distance values surrounding each cell are used to calculate these  parameters. They allow interpolating the DF values for positions within a given cell as:

\begin{equation}
    DF(\mathbf{x})=a_0+a_1 x+a_2 y+ a_3 z + a_4 x y + a_5 x z + a_6 y z + a_7 x y z
\end{equation}

\noindent where $\mathbf{x}=[x, y, z]^t$ represents a 3D point belonging to the grid cell. 

The objective of this interpolation is twofold: First, it allows reducing the impact of the 3D grid space quantization, which may lead to gradient instabilities during the optimization process. Second, and more importantly, this enables analytic gradient computation of the distance with respect to the point under evaluation, accelerating the gradient computation and, hence, the full optimization process.

\reviewLuis{As an alternative option, }\reviewFernando{tricubic interpolation was also implemented and tested into DLL with similar results in terms of accuracy. It was finally discarded in order to reduce the computational requirements and improve the memory usage because it needs 64 float parameters per grid cell. Thus, trilinear interpolation was selected as trade-off between accuracy and computation.}

\subsection{Outlier rejection}

The 3D LIDAR might sense objects that are not mapped (static or dynamic), or some of the points might fall out of the map. We need to deal with these points to prevent degenerated solutions or even divergence in the optimization process. Two main actions are taken to prevent these outliers:

\begin{itemize}
    \item Points out of the map. In this case, the distance function returns a set of interpolation parameters equal to zero. This produce both error and gradient equal to zero, avoiding influencing the optimization process.
    
    \item Unmapped objects. A robust Cauchy Kernel with scaling factor set to $0.1$ is applied to each point of the LIDAR. Thus, when the distance to the closest point is large, the kernel penalizes this constraint reducing its influence into the whole optimization process. \reviewFernando{Huber Kernel was also tested with different scales factors, but the results were less accurate in most cases.}
    
\end{itemize}

\subsection{Implementation details}

\begin{figure*}[!t]
     \centering
     \begin{subfigure}[b]{0.24\textwidth}
         \centering
         \includegraphics[width=\textwidth]{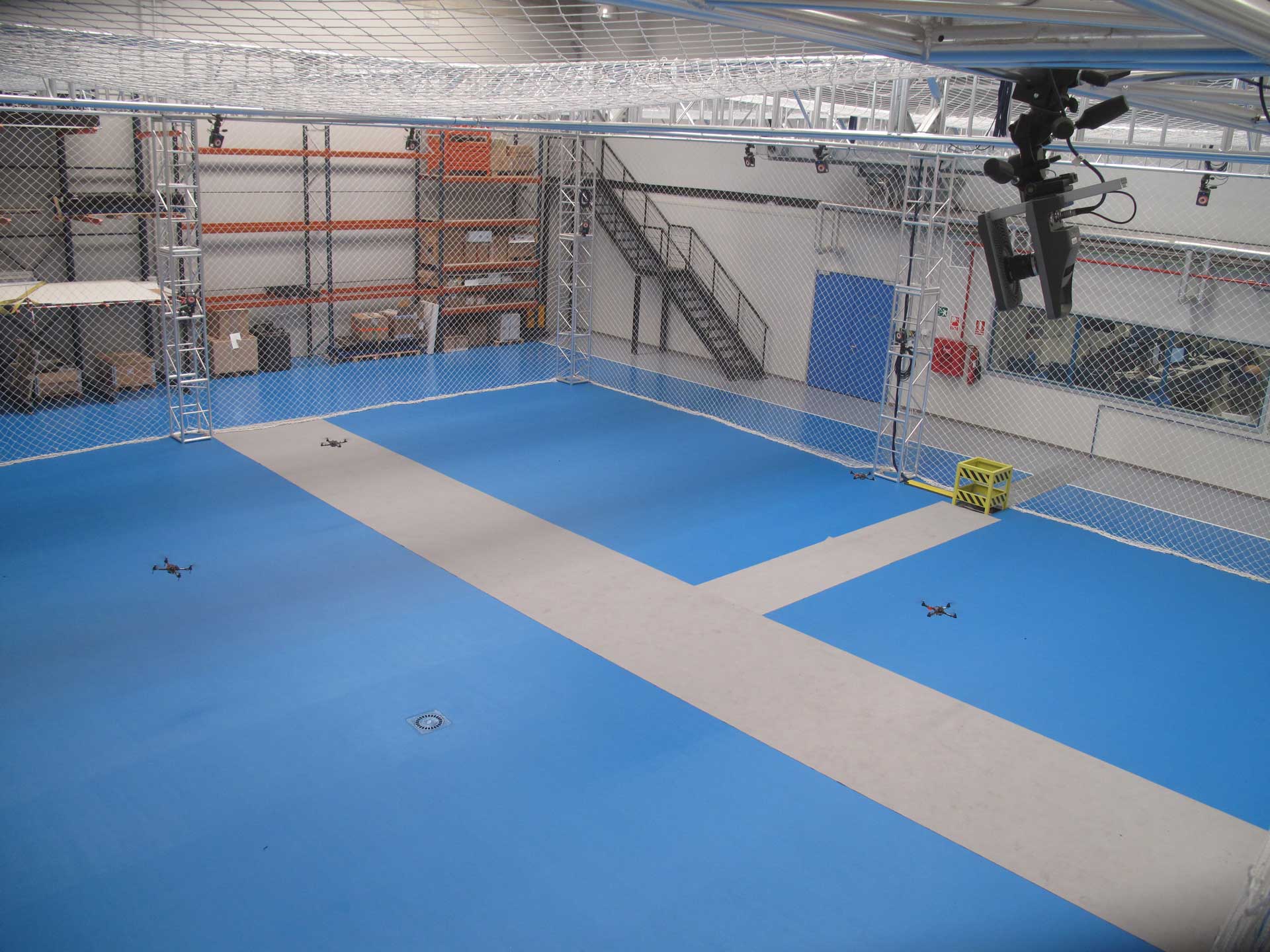}
         \caption{}
         \label{fig:datasets:catec1}
     \end{subfigure}
     \hfill
     \begin{subfigure}[b]{0.24\textwidth}
         \centering
         \includegraphics[width=\textwidth]{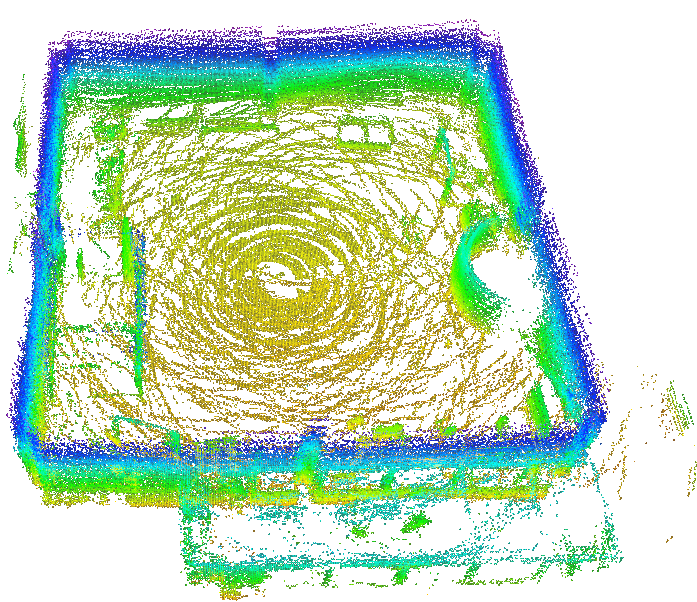}
         \caption{}
         \label{fig:datasets:catec2}
     \end{subfigure}
    \hfill
     \begin{subfigure}[b]{0.24\textwidth}
         \centering
         \includegraphics[width=\textwidth]{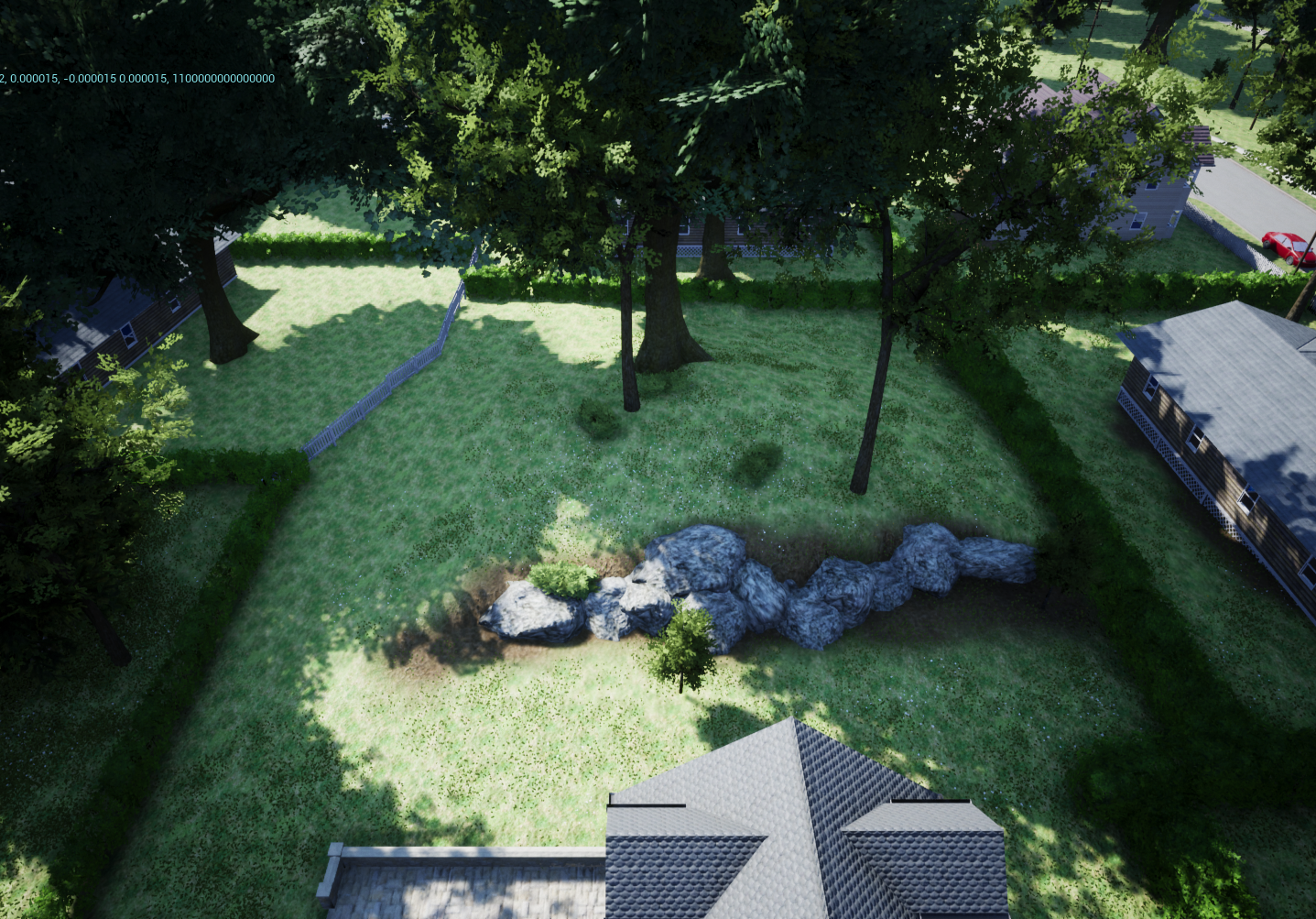}
         \caption{}
         \label{fig:datasets:airsim1}
     \end{subfigure}
     \hfill
     \begin{subfigure}[b]{0.24\textwidth}
         \centering
         \includegraphics[width=0.8\textwidth]{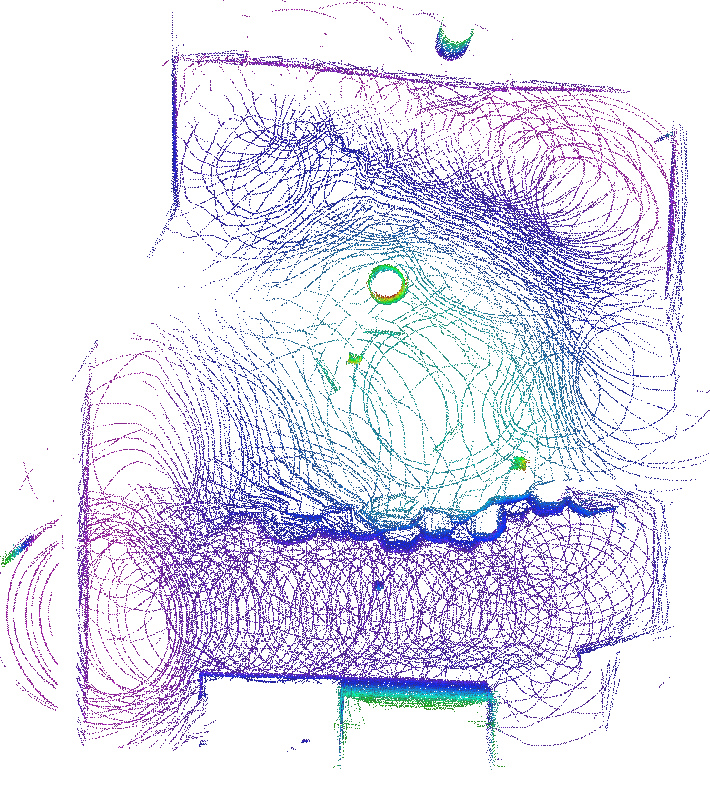}
         \caption{}
         \label{fig:datasets:airsim2}
     \end{subfigure}
    \caption{Catec dataset scenario. (a) Picture of the scenario (source \url{http://www.catec.aero}). (b) 3D map of the scenario, 15x15x5 meters volume. Airsim dataset scenario. (c) Picture of the scenario. (d) 3D map of the scenario, 50x50x10 meters volume.}
    \label{fig:airsimdatasets}
\end{figure*}

DLL has been implemented in C++ and integrated into the ROS framework. We used Ceres Solver \cite{ceres-solver} to implement the non-linear optimization process. The Jacobians have been analytically computed and included into the optimization. 

As previously introduced, DLL makes use of robot odometry and IMU for faster computation. Thus, DLL does not compute roll and pitch angles, the point cloud is tilt compensated before registration instead. Therefore, only X, Y and Z translation, together with yaw angle are estimated during alignment. Please notice that DLL is a general approach, so that estimating roll and pitch is just a matter of adding the corresponding  states and Jacobian into the optimizer.

When a new point cloud must be registered, DLL computes the odometric shift since the last map registration computation and uses it as initial guess for alignment.

%%%%%%%%%%%%%%%%%%%%%%%%%%%%%%%%%%%%%%%%%%%%%%%%%%%%%%%%%%%%%%%%%%%%%%%%%%%%%%%%
\section{Experimental Results}

This section presents the experimental validation of the proposed approach. The method has been tested using a publicly available real indoor dataset, and a simulated outdoor dataset using Airsim \cite{airsim2017fsr}. The approach has been benchmarked against other three methods. The DLL ROS software implementation, and the Airsim simulated dataset can be downloaded here\footnote{\url{https://github.com/robotics-upo/dll}}.

\subsection{Datasets}

We can find a good number of datasets involving ground robots and multiple sensors (cameras, 3D-LIDAR, RADAR, etc) in the state of the art. However, the number of datasets involving aerial robots and 3D LIDAR is limited in the literature. This is mainly motivated by the weight constraints of these systems.

\begin{figure*}[!t]
     \centering
     \begin{subfigure}[b]{0.24\textwidth}
         \centering
         \includegraphics[width=\textwidth]{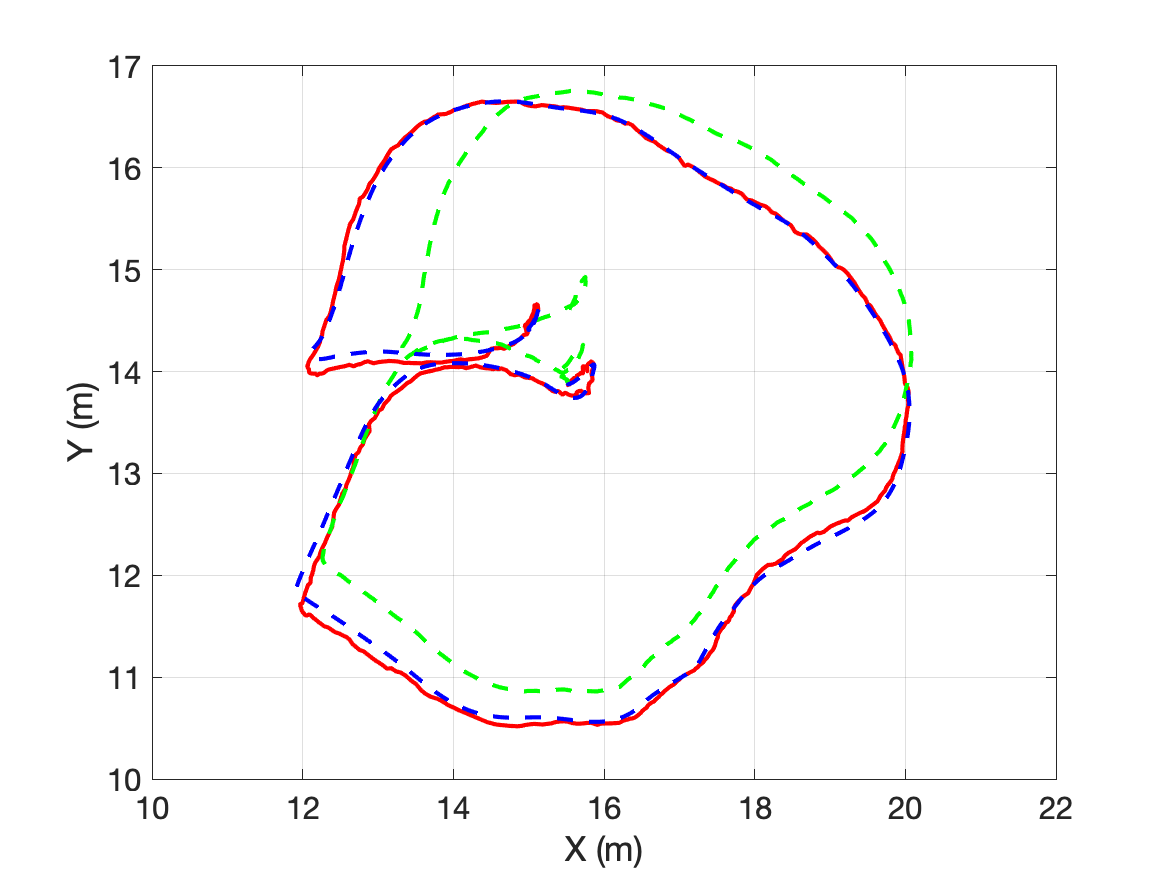}\\
         \includegraphics[width=\textwidth]{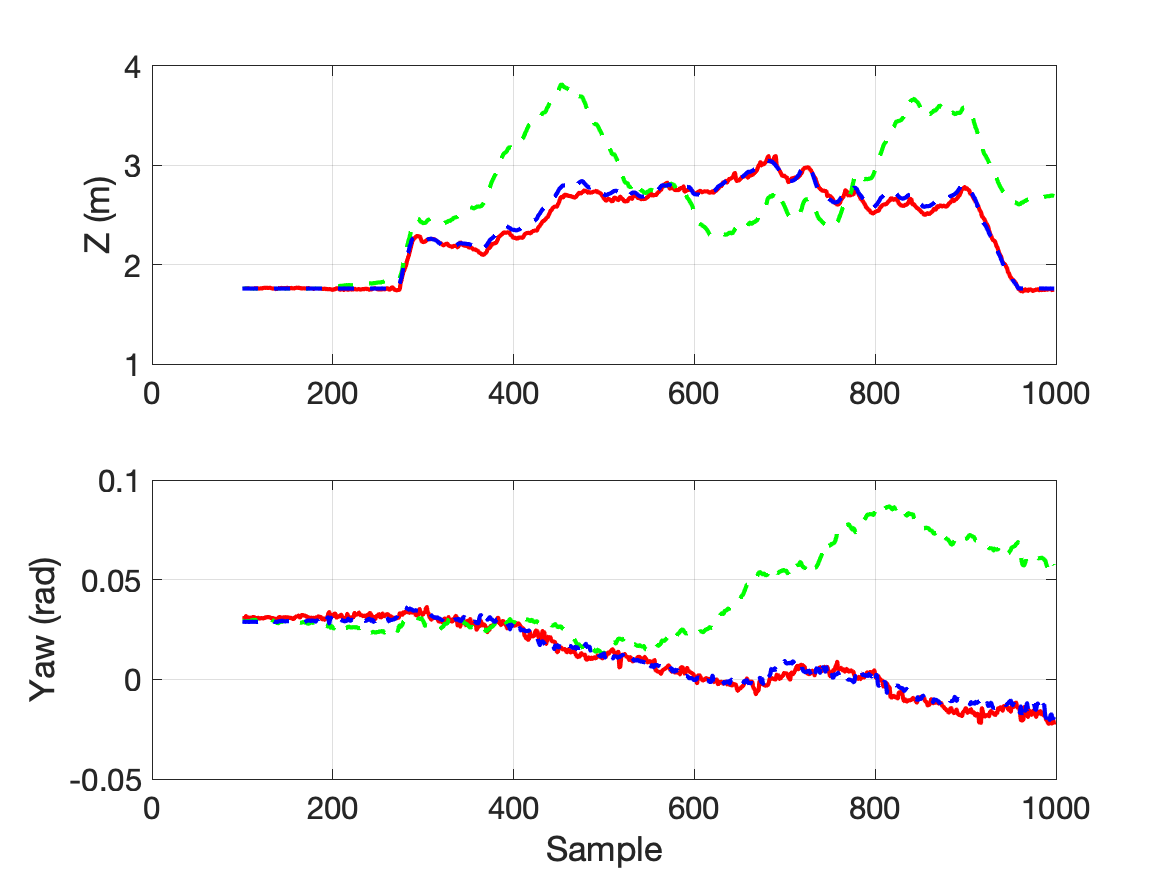}
         \caption{}
         \label{fig:trajs:catec1}
     \end{subfigure}
     \hfill
     \begin{subfigure}[b]{0.24\textwidth}
         \centering
         \includegraphics[width=\textwidth]{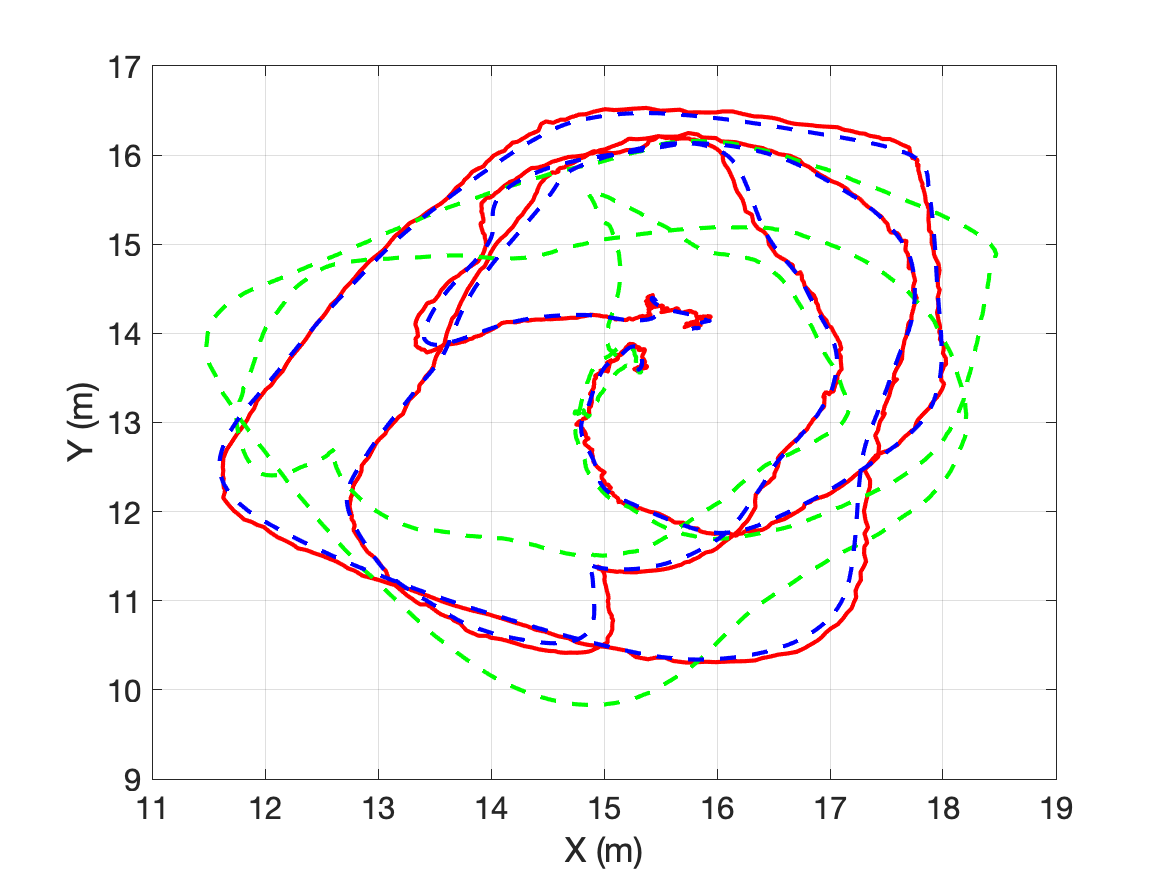}\\
         \includegraphics[width=\textwidth]{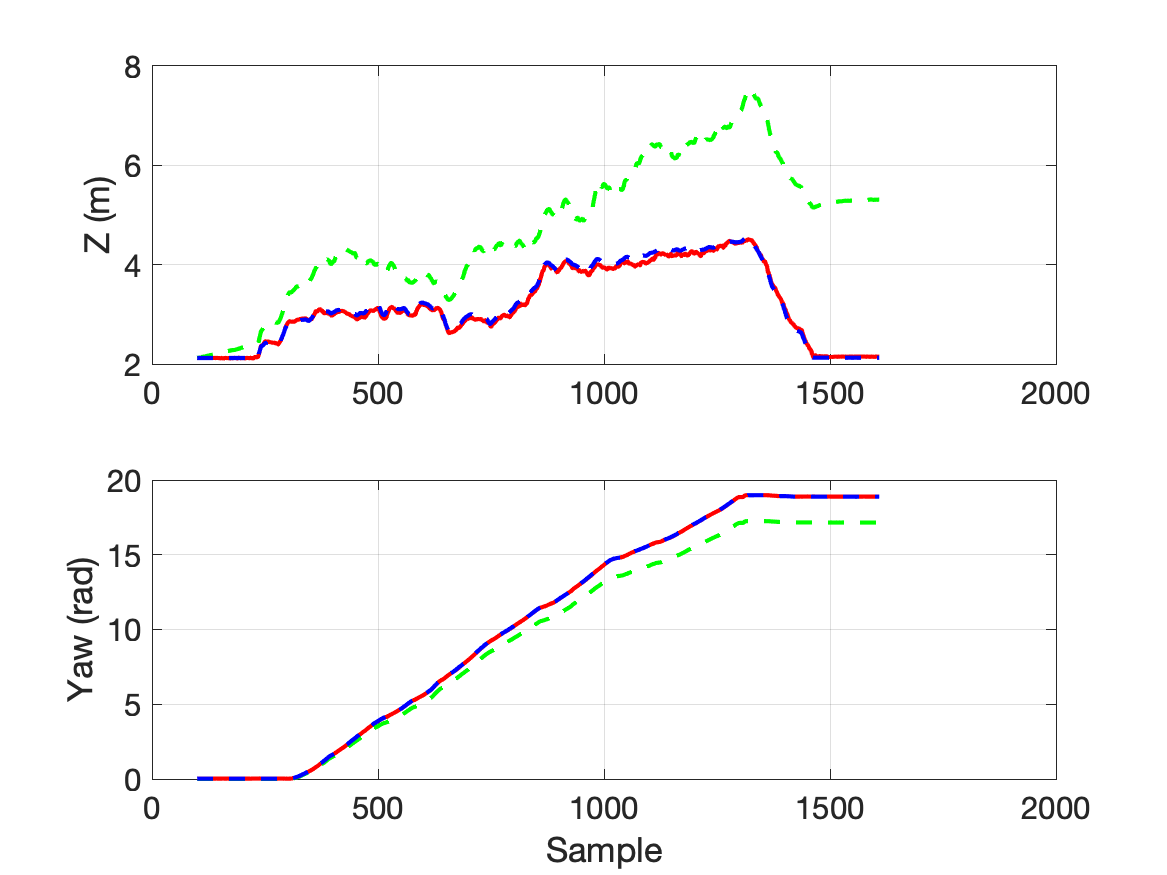}
         \caption{}
         \label{fig:trajs:catec2}
     \end{subfigure}
     \hfill
     \begin{subfigure}[b]{0.24\textwidth}
         \centering
         \includegraphics[width=\textwidth]{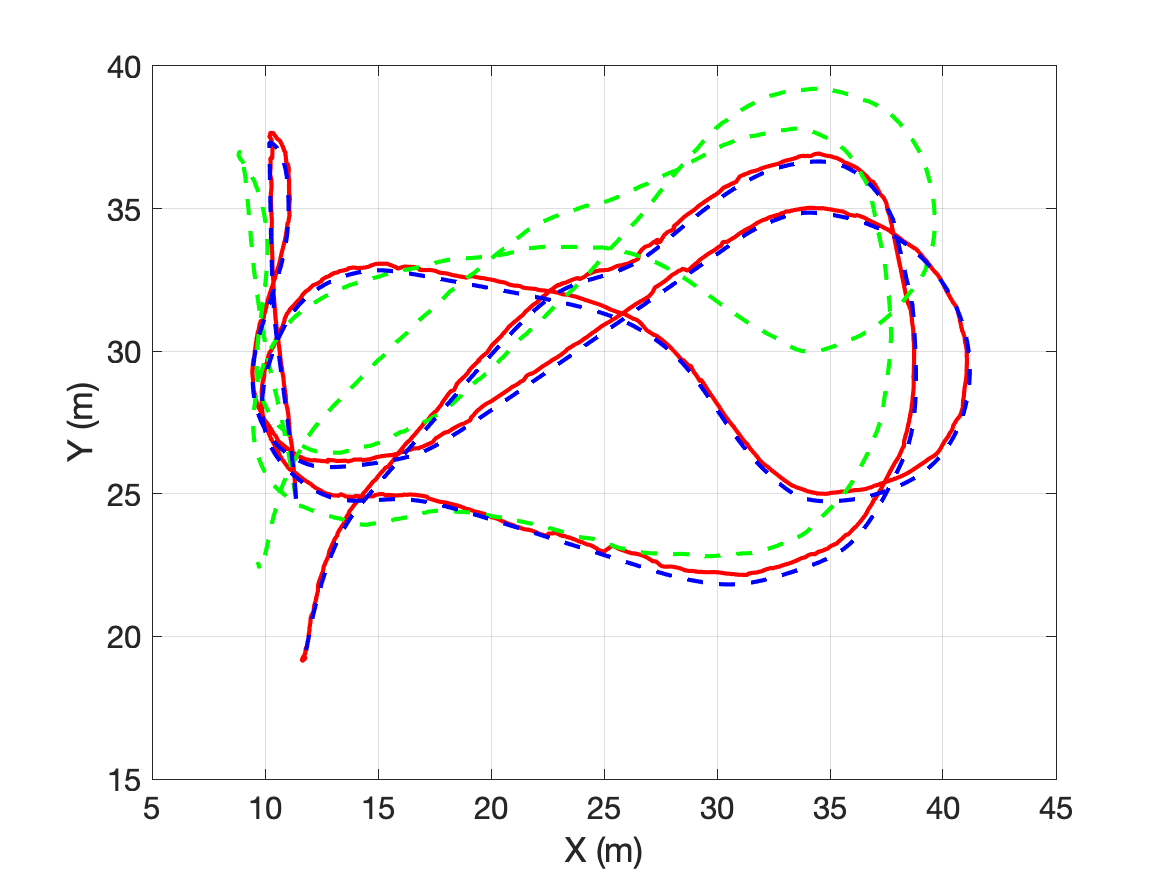}\\
         \includegraphics[width=\textwidth]{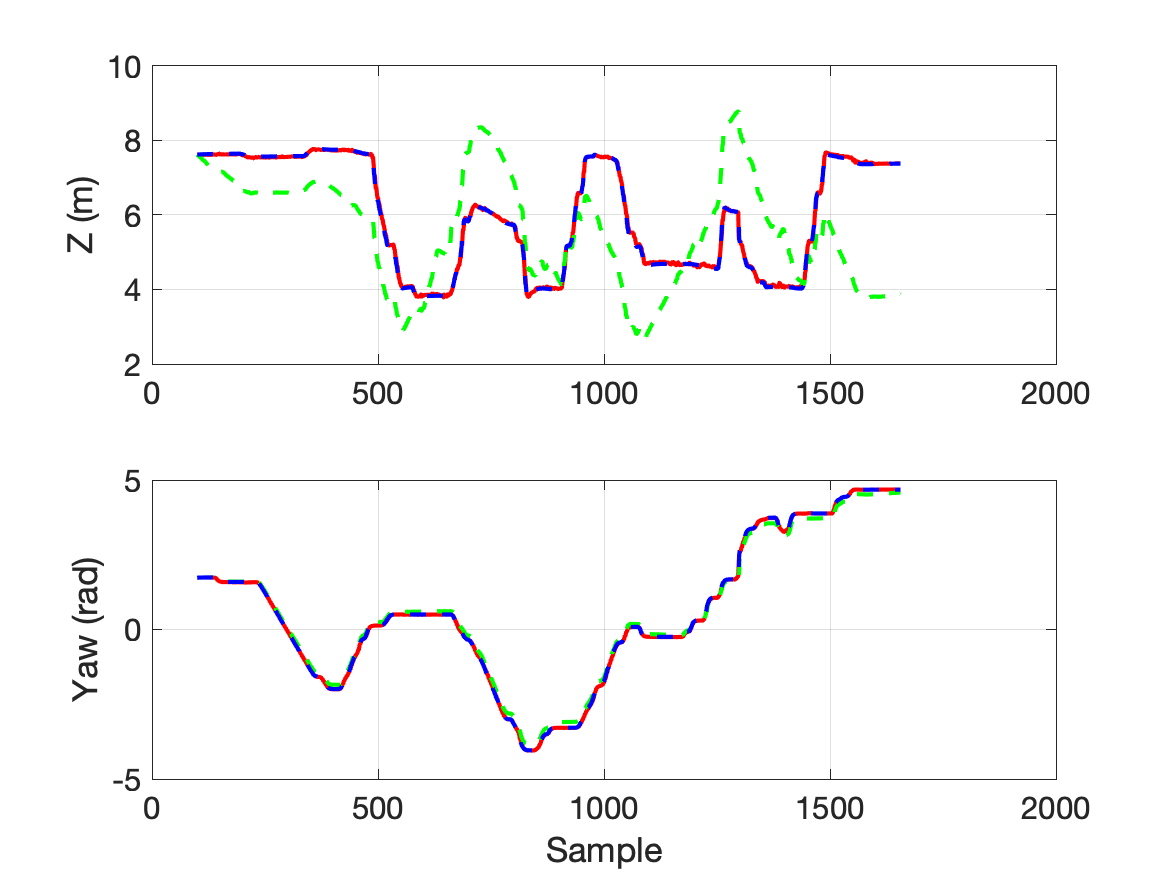}
         \caption{}
         \label{fig:trajs:airsim1}
     \end{subfigure}
     \hfill
     \begin{subfigure}[b]{0.24\textwidth}
         \centering
         \includegraphics[width=\textwidth]{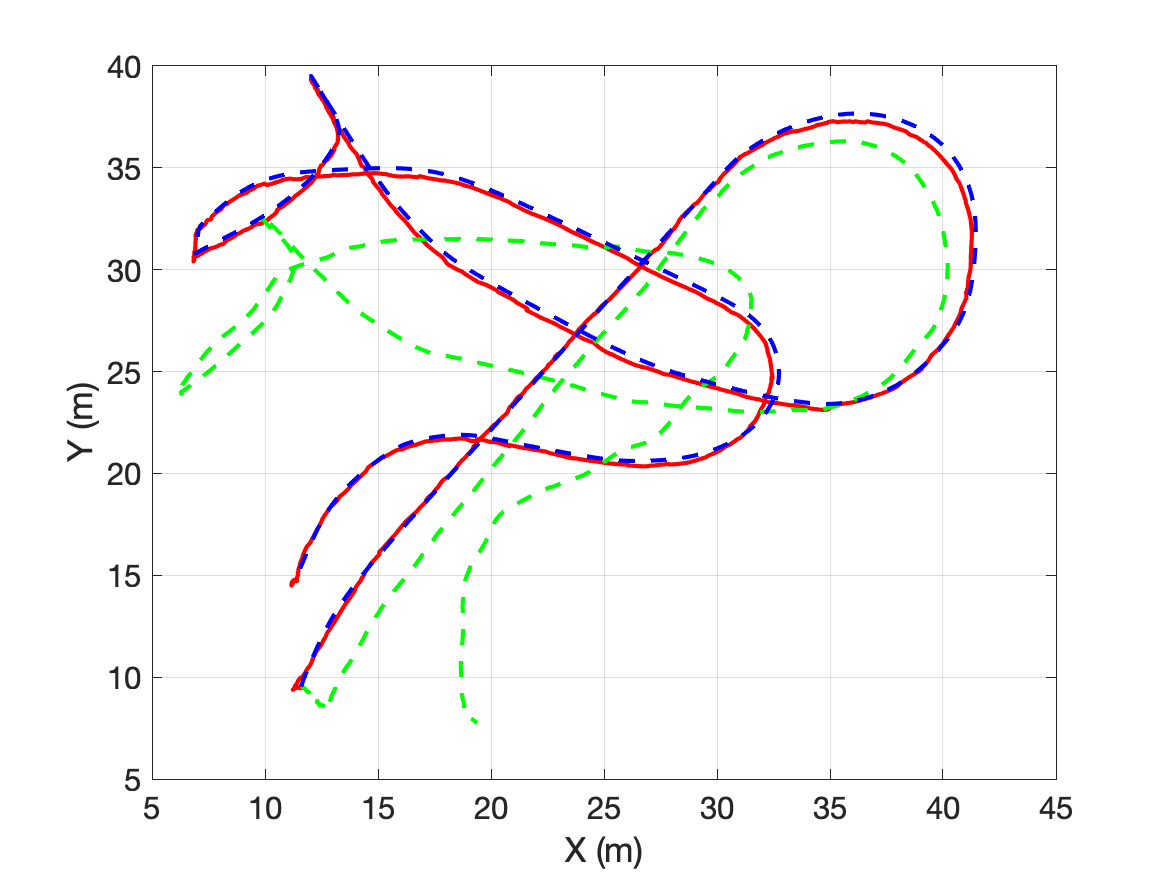}\\
         \includegraphics[width=\textwidth]{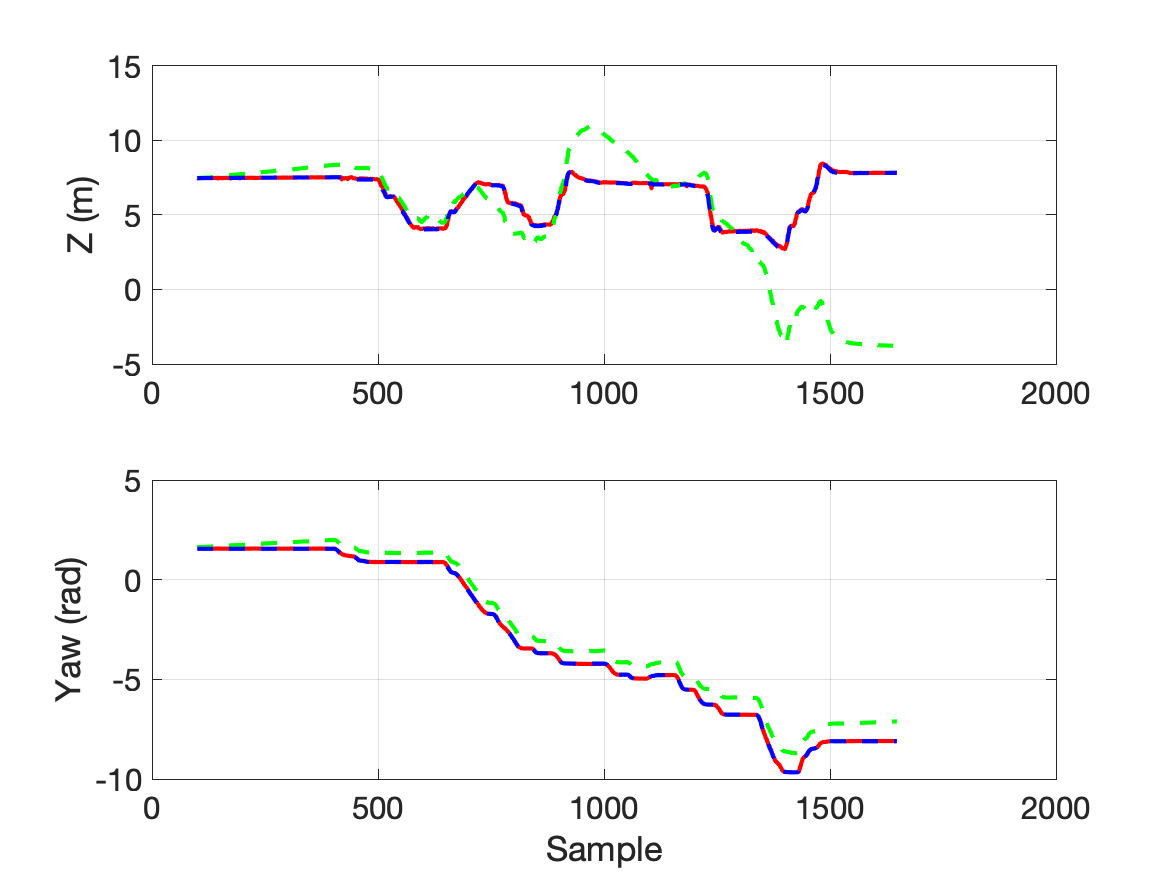}
         \caption{}
         \label{fig:trajs:airsim2}
     \end{subfigure}
    \caption{Dataset ground-truth trajectories (dashed blue line), DLL estimations (solid red line) and odometry estimation (dashed green line). Top: X-Y view. Middle: Z coordinate. Bottom: Yaw angle evolution. (a) \texttt{catec1} trajectory. (b) \texttt{catec2} trajectory. (c) \texttt{airsim1} trajectory. (d) \texttt{airsim2} trajectory.}
    \label{fig:trajs}
\end{figure*}

%We will validate the proposed system in two different datasets, one real indoor dataset, and a simulated outdoor dataset. 
The indoor dataset is composed by two fights released together with the \emph{AMCL3D ROS}\footnote{\url{http://wiki.ros.org/amcl3d}} module \cite{Perezgrau17JARS} for aerial robot localization indoors\footnote{The dataset can be downloaded from \url{https://github.com/fada-catec/amcl3d/releases}}. The scenario is shown in Fig. \ref{fig:datasets:catec1}, and consists of a $15\times15\times5$ meters indoors volume. The XY trajectory of both flights are shown in Figs. \ref{fig:trajs:catec1} and \ref{fig:trajs:catec2}. The \texttt{catec1} trajectory is $31.40$ meter-long, while \texttt{catec2} is $56.69$ meters. The dataset contains the ground-truth robot position provided by a VICON motion capture system with millimeter accuracy, and the 3D LIDAR data of a 16 channel sensor attached to the vehicle.  

In order to validate the method in a different environment, a dataset using Airsim \cite{airsim2017fsr} has been created (which can be downloaded together with DLL code). Airsim provides simulated realistic images and sensors from both aerial and ground robots. The scenario is a $50\times50\times10$ meters volume, shown in Fig. \ref{fig:datasets:airsim1}. It contains two flights of a simulated drone with a 16 channel 3D LIDAR attached. Figure \ref{fig:trajs:airsim1} shows the XY trajectory performed by the drone in the \texttt{airsim1} flight, and Fig. \ref{fig:trajs:airsim2} the corresponding trajectory of the \texttt{airsim2} flight. The total length of the trajectories are $191.94$ and $164.92$ meters respectively. The ground-truth in this case is provided by the simulator, so we assume it is error free.

\reviewFernando{It is worth to mention that the \texttt{airsim} scenarios cannot be sensed completely by a single LIDAR scan, as it does in \texttt{catec} datasets. This is made on purpose in order to properly validate the localization approach in a more usual setup in which the environment cannot be seen/sensed at once due to its size or because of occlusions. Also, the real scenario used in Section \ref{sec:mbzirc} is large and cannot be sensed at once by the LIDAR.}

%\begin{figure}[!t]
%     \centering
%     \begin{subfigure}[b]{0.48\columnwidth}
%         \centering
%         \includegraphics[width=\columnwidth]{figures/catec_view.jpg}
%         \caption{}
%         \label{fig:datasets:catec}
%     \end{subfigure}
%     \hfill
%     \begin{subfigure}[b]{0.48\columnwidth}
%         \centering
%         \includegraphics[width=\columnwidth]{figures/catec.png}
%         \caption{}
%         \label{fig:datasets:airsim}
%     \end{subfigure}
    
%    \caption{Catec dataset scenario. (a) Picture of the scenario (source \url{http://www.catec.aero}). (b) 3D map of the scenario, 15x15x5 meters volume.}
%    \label{fig:catecdatasets}
%\end{figure}

%\begin{figure}[!t]
%     \centering
%     \begin{subfigure}[b]{0.48\columnwidth}
%         \centering
%         \includegraphics[width=\columnwidth]{figures/airsim_view.png}
%         \caption{}
%         \label{fig:datasets:catec}
%     \end{subfigure}
%     \hfill
%     \begin{subfigure}[b]{0.48\columnwidth}
%         \centering
%         \includegraphics[width=\columnwidth]{figures/airsim.png}
%         \caption{}
%        \label{fig:datasets:airsim}
%     \end{subfigure}
    
%    \caption{Airsim dataset scenario. (a) Picture of the scenario. (b) 3D map of the scenario, 50x50x10 meters volume.}
%    \label{fig:airsimdatasets}
%\end{figure}

\subsection{Benchmarks}

Three other localization solutions have been tested in order to put in context the localization results of DLL:
\begin{itemize}
    \item ICP: an ICP-based localization system has been implemented using the ICP solution provided by the PCL library \cite{Rusu_ICRA2011_PCL}. It was setup with $50$ iterations, $0.1$ meters of max correspondence distance and outlier rejection with $1.0$ meters threshold. The target cloud for ICP (the map) was computed offline and not included into the processing time computation. This method has been integrated into the DLL ROS module as a different point cloud registration mechanism, and it is publicly available together with the rest of the source code.
    
    \item NDT: an NDT-based \cite{magnusson2007scan} localization system has been implemented using the solution provided by the PCL library \cite{Rusu_ICRA2011_PCL}. It was setup with $50$ iterations, $1.0$ meters resolution of the NDT grid, $0.01$ meter of transformation epsilon and $0.1$ meter for the maximum step size for the More-Thuente line search. The NDT representation of the map was computed offline and not included into the processing time computation. This method has been integrated into the DLL ROS module as a different point cloud registration mechanism, and it is publicly available together with the rest of the source code.
    
    \item AMCL3D: the Montecarlo Localization approach implemented in \emph{AMCL3D ROS} has been used. This code implements the method described in \cite{Perezgrau17JARS}. It was configured with $1000$ particles, $0.2$ meters/radians for additive noise in prediction function for X, Y, Z and yaw, and $0$ resample interval. These are the parameters provided for \texttt{catec1} and \texttt{catec2} trajectories in the paper. 
    
\end{itemize}

As mentioned in Section \ref{sec:method}, DLL (and most localization approaches) makes use of the robot odometry to obtain a prior about the robot position into the map. For these experiments, the LiDAR odometry and mapping method LOAM\footnote{\url{https://github.com/HKUST-Aerial-Robotics/A-LOAM}} was used to compute the odometry. The method was configured to perform only odometry, the mapping has been deactivated to prevent jumps in the LiDAR-to-LiDAR pose estimation. This odometry was computed for each of the datasets, and served to each of the methods above mentioned as initial solution.  

The \reviewFernando{map resolution (map point cloud and distance field)} in both scenarios is the same, $0.05$ meters. The point-cloud size and density of the 3D LIDAR was also the same for all the approaches, except for NDT. As pointed out in \cite{Akai2020}, we needed to apply a $2$ meters voxel grid filter to the LIDAR point cloud, otherwise the PCL NDT implementation did not converge to a solution. 

Finally, all the implementations make use of LIDAR tilt-compensation based on the onboard IMU. Thus, the point clouds used for registration are already compensated according to the roll and pitch angles provided by the drone IMU.

All the tests were carried out in the same computer, a 10$^{th}$ generation Intel Core i7 laptop with 16 GB of RAM running Ubuntu 18.04. Although the Ceres Solver can be configured to run in multiple threads, all the methods have been configured to be single threaded in order to have a common baseline. 

The trajectory computed by DLL is presented together with the ground-truth and the odometry in Fig \ref{fig:trajs}. The results of the benchmarks are shown in Table \ref{tab:bench}. The table shows the computed RMSE in position and yaw angle. It also shows the averaged processing time per point cloud $\Delta t$. It can be seen how the computed errors in DLL are small, in the same order of NDT and ICP, and for some experiments DLL provides the smallest error. It is also worth to mention that the computation time per point-cloud is about one order of magnitude smaller than all the other approaches. 

Thus, according to the benchmark results, DLL presents an accuracy at the same level of other approaches in the state of the art, and even better with very small impact in computation. Even more, the DLL computational time can be improved about $30\%$ just increasing the number of threads used by the Ceres Solver.

\begin{table*}[!t]
    \centering
    \begin{tabular}{|l l|c|c|c|c|}
\hline
  Scenario	&	&DLL  	&ICP  	&NDT   	&AMCL3D\\
\hline
\hline
\texttt{catec1}  &rmse$_t$ (dev)   &\bf{0.0548 (0.0556)} &0.0578 (0.0619) &0.0602 (0.0638) &0.1674 (0.1828)\\
        &rmse$_a$ (dev)   &0.0030 (0.0018) &\bf{0.0025 (0.0016)} &0.0044 (0.0025) &0.0271 (0.0033)\\
	    &$\Delta t$ (dev) &\bf{0.0731 (0.0237)} &2.1327 (0.1231) &0.4917 (0.0616) &0.5367 (0.0621)\\
\hline
\texttt{catec2}	&rmse$_t$ (dev)   &0.0580 (0.0480) &\bf{0.0571 (0.0554)} &0.0605 (0.0542) &0.8715 (1.0778)\\ 
	    &rmse$_a$ (dev)   &0.0116 (0.0067) &0.0185 (0.0117) &\bf{0.0077 (0.0046)} &0.7512 (0.4854)\\
	    &$\Delta t$ (dev) &\bf{0.0763 (0.0242 )} &2.2236 (0.1470) &0.6299 (0.1420) &0.5702 (0.0674)\\
\hline
\texttt{airsim1} &rmse$_t$ (dev)   &0.0920 (0.0795) &0.1161 (0.0805) &\bf{0.0696 (0.0692)} &1.3302 (12.4577)\\
	    &rmse$_a$ (dev)   &\bf{0.0181 (0.0137)} &0.0211 (0.0170) &0.0185 (0.0151) &0.1142 (0.1047)\\
	    &$\Delta t$ (dev)   &\bf{0.0370 (0.0141)} &0.7908 (0.1630) &0.5211 (0.3055) &0.3686 (0.1270)\\
\hline
\texttt{airsim2} &rmse$_t$ (dev)   &0.1027 (0.1445) &0.0753 (0.0878) &\bf{0.0719 (0.0815)} &2.9354 (9.8130)\\
	    &rmse$_a$ (dev)   &\bf{0.0142 (0.0106)}	&0.0165 (0.0146)	&0.0154	(0.0135)&0.7407 (0.0978)\\
        &$\Delta t$ (dev) &\bf{0.0372 (0.0135)} &0.7917 (0.1408) &0.5427 (0.3855) &0.3942 (0.0915)\\
\hline
    \end{tabular}
    \caption{Localization benchmark results. rmse$_t$ in meters, rmse$_a$ in radians and $\Delta t$ in seconds.}
    \label{tab:bench}
\end{table*}

\subsection{Robustness study}

Most localization approaches, direct or feature-based, work well and efficiently when the robot odometry is accurate. A good prior about the robot position helps to reduce the searching space for the solution, impacting both accuracy and computational time. However, in real-world applications the odometry estimation might be temporally noisy or even jumpy. This section will benchmark the proposed approach focusing in robustness.

Thus, the following experiment has been conceived to evaluate the robustness of the approach against noisy odometry: the four algorithms have been run over the same experiments under three different setups. Particularly, the \texttt{catec1} trajectory has been selected. In each trial, random noise has been added to the translation and yaw angle increment computed from odometry. The following setups have been implemented:
\begin{itemize}
    \item \texttt{Baseline}: This is the setup presented in Table \ref{tab:bench} in scenario \texttt{catec1}.
    \item \texttt{NoOdom}: In this setup, no odometry is used for localization. The filters are initialized in the correct position. Then, in each filter iteration, they use the previous solution as initial guess for the new update.
    \item \texttt{MidNoise}: In this setup, Gaussian noise is added to the translation and yaw angle increments computed from odometry. This noise is centered in the actual translation/rotation computed from odometry and it has a standard deviation of $0.25$ meters in each axis and $0.05$ radians in yaw.
    \item \texttt{LargeNoise}: Similar to \texttt{MidNoise} setup, but with a standard deviation of $0.5$ meters in each axis and $0.1$ radians in yaw.
\end{itemize}

Table \ref{tab:robustness} summarizes the errors and computation time of the different approaches in the setup previously described. It can be seen how DLL provides the smallest error in many cases, and also provides errors very close to the best solution in the rest of the cases. It is worth to mention that DLL and NDT are the only approaches able to stand with mid and large errors in odometry. Notice that DLL and NDT were able to compute a solution even with jumps of $0.75$ meters and $0.12$ radians in the odometry. Among all the approaches, DLL provides the fastest solution, roughly an order of magnitude faster.

\begin{table*}[!t]
    \centering
    \begin{tabular}{|l l|c|c|c|c|}
\hline
  Setup	&	               &DLL  	&ICP  	&NDT   	&AMCL3D\\
\hline
\hline
\texttt{Baseline} & rmse$_t$ (dev)     &\bf{0.0548} (0.0556) &0.0578 (0.0619) &0.0602 (0.0638) &0.1674 (0.1828) \\
         & rmse$_a$ (dev)     &0.0030 (0.0018) &\bf{0.0025} (0.0016) &0.0044 (0.0025) &0.0271 (0.0033) \\
         & $\Delta t$ (dev) &\bf{0.0731} (0.0237) &2.1327 (0.1231) &0.4917 (0.0616) &0.5367 (0.0621) \\
\hline   
\texttt{NoOdom} & rmse$_t$ (dev)       &0.0766 (0.0525) &\bf{0.0612} (0.0494) &0.0838 (0.0648) &0.4242 (0.2346) \\
         & rmse$_a$ (dev)     &0.0032 (0.0019) &\bf{0.0030} (0.0016) &0.0049 (0.0027) &0.1109 (0.0376) \\
         & $\Delta t$ (dev) &\bf{0.0903} (0.0268) &2.1954 (0.0952) &0.4958 (0.0651) &0.4535 (0.1072) \\
\hline
\texttt{MidNoise} & rmse$_t$ (dev)     &\bf{0.0899} (0.0547) &* &0.1061 (0.0761) &* \\
         & rmse$_a$ (dev)     &0.0050 (0.0032) &* &\bf{0.0048} (0.0027) &* \\
         & $\Delta t$ (dev) &\bf{0.1168} (0.0260) &* &1.2133 (0.4142) &* \\
\hline
\texttt{LargeNoise} & rmse$_t$ (dev)   &\bf{0.1457} (0.1923) &* &0.4951 (0.6200) &* \\
         & rmse$_a$ (dev)     &\bf{0.0103} (0.0092) &* &0.1337 (0.1223) &* \\
         & $\Delta t$ (dev) &\bf{0.1381} (0.0313) &* &1.9501 (0.7216) &* \\
\hline
    \end{tabular}
    \caption{Robustness benchmark results. rmse$_t$ in meters, rmse$_a$ in radians and $\Delta t$ in seconds. Entries with * indicate that the method diverged during the execution in such setup.}
    \label{tab:robustness}
\end{table*}

\subsection{Validation in MBZIRC2020 dataset}
\label{sec:mbzirc}
The approach has been also validated using data from a real outdoor flight carried out by our SkyEye Team during the Mohamed Bin Zayed International Robotics Challenge 2020, Challenge 3\footnote{\url{http://www.mbzirc.com/challenge/2020}}. This experiment was not used as benchmark because the available reference is based on GPS in single configuration, so the accuracy was not enough to guarantee a fair comparison among approaches. However, it can be used to show that DLL is able to also perform well outdoors in a large environment.

\begin{figure}
    \centering
    \includegraphics[width=0.8\columnwidth]{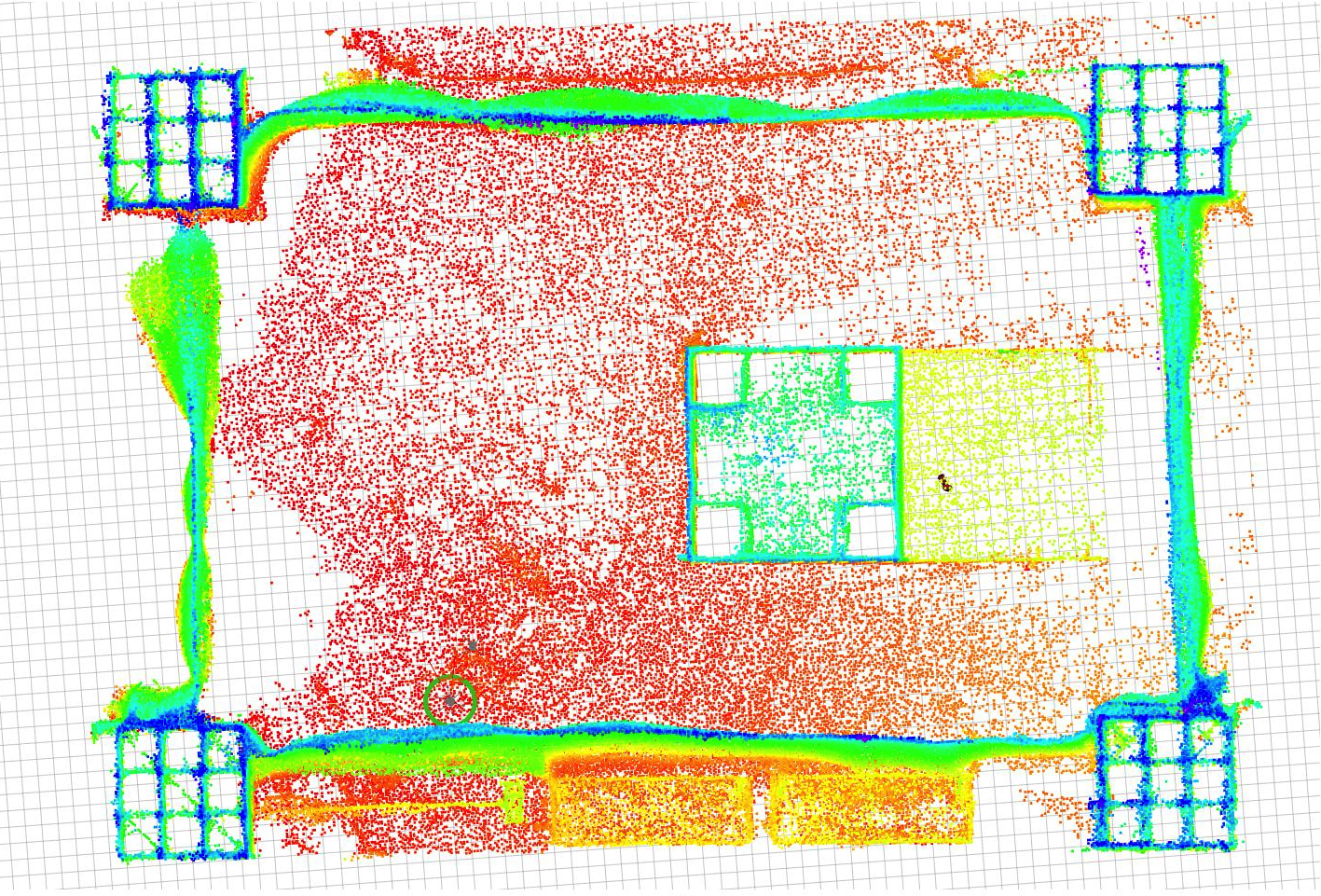}
    \caption{Top view of the 3D map of the MBZIRC 2020 real outdoor dataset. \reviewFernando{The colors indicate the altitude of the points: red (low altitute, around 0 meters), green (mid altitude, about 15 meters) and blue (hight altitude, about 30 meters).}}
    \label{fig:mbzirc-dataset}
\end{figure}

Figures \ref{fig:mbzirc-arena} and \ref{fig:mbzirc-dataset} show the  environment where the dataset was gathered. It is a volume of $60\times 30\times 20$ meters approximately. It is composed by a $15$ meters high building in the center of an arena surrounded by four towers. 

\begin{figure}
    \centering
    \includegraphics[width=\columnwidth]{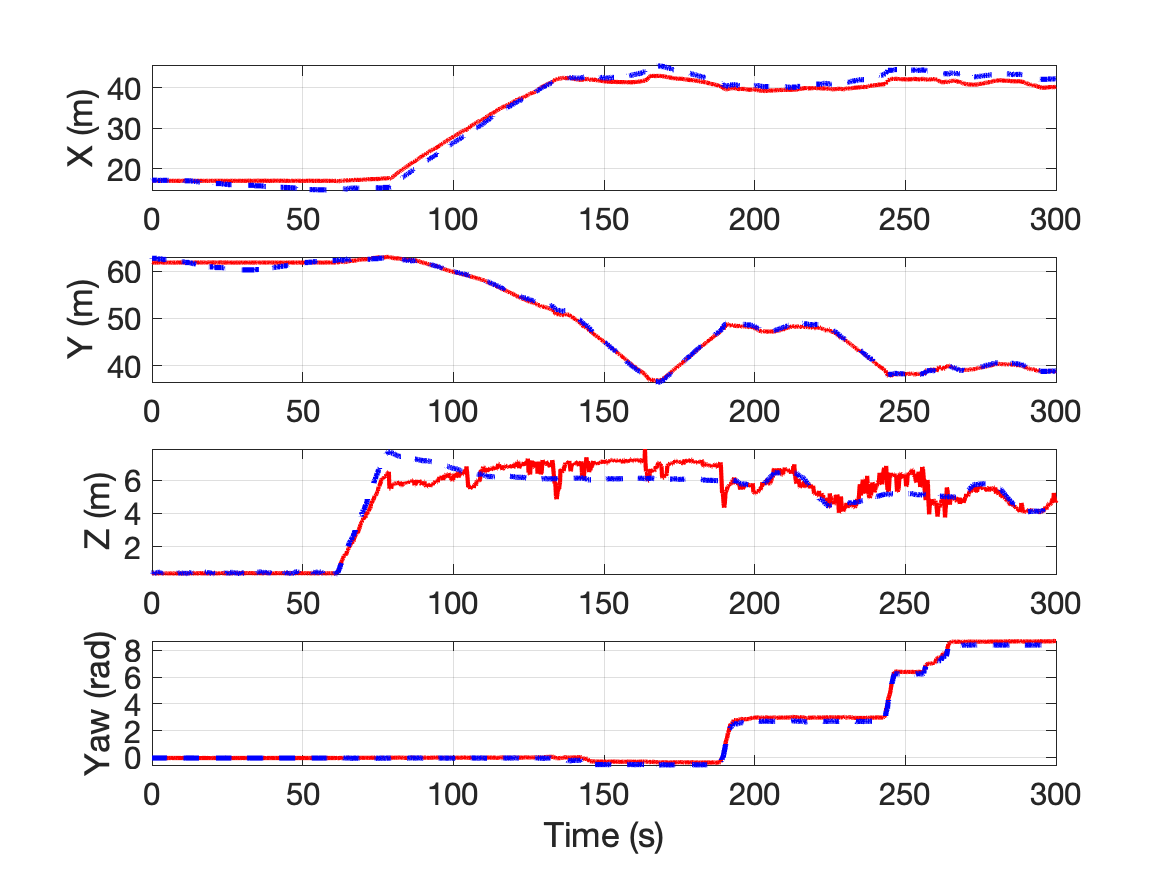}
    \caption{Estimated UAV position in MBZIRC2020 dataset. GPS baseline (dashed blue line) and   DLL estimation (solid red line).}
    \label{fig:mbzirc-plot}
\end{figure}

The estimation by DLL compared to the baseline is presented in Fig. \ref{fig:mbzirc-plot}. The baseline XY position is provided by the UAV (DJI's M210v2) GPS in single configuration, the Z reference is provided by a barometric altimeter and the yaw angle is compared against the UAV compass. It can be seen how the DLL estimation closely follows the baseline. 

%%%%%%%%%%%%%%%%%%%%%%%%%%%%%%%%%%%%%%%%%%%%%%%%%%%%%%%%%%%%%%%%%%%%%%%%%%%%%%%%
\section{Conclusions}
The paper presents a direct map-based pose tracking approach using 3D LIDAR. The method uses the raw point clouds and avoids the search for point correspondences. The predicted pose from odometry is refined by directly optimizing the distance of the points to the map, using a distance field representation of the latter. 

The results show how the approach achieves similar precision as competing methods like NDT and ICP for the scenarios considered, but runs one order of magnitude faster, allowing real-time execution. Furthermore, the optimization-based approaches NDT and DLL behave better than the Monte-Carlo localization approach in the case of noisy odometry, following also the findings of \cite{dantanarayana2016c}. DLL is able to handle larger noise than NDT, achieving lower errors in this case.

Future work will consider improving the space efficiency of the method. One of the main drawbacks of the proposed method is the amount of memory required by the 3D grid representation of the distance field. More efficient structures for the distance field like in \cite{Akai2020} will be considered. \reviewFernando{Future work will also study possible strategies to estimate the uncertainty associated to the computed solution. This information can be used by others algorithms (like planners or controllers) as quantitative criteria to take decisions related with the localization estimation.}

%%%%%%%%%%%%%%%%%%%%%%%%%%%%%%%%%%%%%%%%%%%%%%%%%%%%%%%%%%%%%%%%%%%%%%%%%%%%%%%%
%\section*{Acknowledgment}

%We thank ...

\balance
\bibliographystyle{IEEEtran}
\bibliography{bibliography}

\begin{thebibliography}{10}
\providecommand{\url}[1]{#1}
\csname url@rmstyle\endcsname
\providecommand{\newblock}{\relax}
\providecommand{\bibinfo}[2]{#2}
\providecommand\BIBentrySTDinterwordspacing{\spaceskip=0pt\relax}
\providecommand\BIBentryALTinterwordstretchfactor{4}
\providecommand\BIBentryALTinterwordspacing{\spaceskip=\fontdimen2\font plus
\BIBentryALTinterwordstretchfactor\fontdimen3\font minus
  \fontdimen4\font\relax}
\providecommand\BIBforeignlanguage[2]{{%
\expandafter\ifx\csname l@#1\endcsname\relax
\typeout{** WARNING: IEEEtran.bst: No hyphenation pattern has been}%
\typeout{** loaded for the language `#1'. Using the pattern for}%
\typeout{** the default language instead.}%
\else
\language=\csname l@#1\endcsname
\fi
#2}}

\bibitem{rs11111348}
H.~Liu, Q.~Ye, H.~Wang, L.~Chen, and J.~Yang, ``A precise and robust
  segmentation-based lidar localization system for automated urban driving,''
  \emph{Remote Sensing}, vol.~11, no.~11, 2019.

\bibitem{LOAM}
Zhang and S.~Singh, ``{LOAM: Lidar Odometry and Mapping in Real-time},'' in
  \emph{Robotics: Science and Systems Conference (RSS)}, 2014.

\bibitem{ulas13}
C.~Ulas and H.~Temeltaş, ``A fast and robust feature-based scan-matching
  method in 3d slam and the effect of sampling strategies,''
  \emph{International Journal of Advanced Robotic Systems}, vol.~10, 2013.

\bibitem{FCGF2019}
C.~Choy, J.~Park, and V.~Koltun, ``Fully convolutional geometric features,'' in
  \emph{International Conference on Computer Vision - ICCV}, 2019.

\bibitem{gojcic20193DSmoothNet}
Z.~Gojcic, C.~Zhou, J.~D. Wegner, and W.~Andreas, ``The perfect match: 3d point
  cloud matching with smoothed densities,'' in \emph{International conference
  on computer vision and pattern recognition (CVPR)}, 2019.

\bibitem{ChenM92}
Y.~Chen and G.~G. Medioni, ``Object modelling by registration of multiple range
  images.'' \emph{Image Vis. Comput.}, vol.~10, no.~3, pp. 145--155, 1992.

\bibitem{magnusson2007scan}
M.~Magnusson, A.~Lilienthal, and T.~Duckett, ``Scan registration for autonomous
  mining vehicles using 3d-ndt,'' \emph{Journal of Field Robotics}, vol.~24,
  no.~10, pp. 803--827, 2007.

\bibitem{8206170}
H.~{Hong} and B.~H. {Lee}, ``Probabilistic normal distributions transform
  representation for accurate 3d point cloud registration,'' in \emph{2017
  IEEE/RSJ International Conference on Intelligent Robots and Systems (IROS)},
  2017, pp. 3333--3338.

\bibitem{jones20063d}
M.~W. Jones, J.~A. Baerentzen, and M.~Sramek, ``3d distance fields: A survey of
  techniques and applications,'' \emph{IEEE Transactions on visualization and
  Computer Graphics}, vol.~12, no.~4, pp. 581--599, 2006.

\bibitem{elhousni2020survey}
M.~Elhousni and X.~Huang, ``A survey on 3d lidar localization for autonomous
  vehicles,'' in \emph{2020 IEEE Intelligent Vehicles Symposium (IV)}.\hskip
  1em plus 0.5em minus 0.4em\relax IEEE, pp. 1879--1884.

\bibitem{zhang2017low-drift}
J.~Zhang and S.~Singh, ``Low-drift and real-time lidar odometry and mapping,''
  \emph{Auton. Robots}, pp. 401--416, 2017.

\bibitem{dube2018incremental}
R.~Dub{\'e}, M.~G. Gollub, H.~Sommer, I.~Gilitschenski, R.~Siegwart, C.~Cadena,
  and J.~Nieto, ``Incremental-segment-based localization in 3-d point clouds,''
  \emph{IEEE Robotics and Automation Letters}, vol.~3, no.~3, pp. 1832--1839,
  2018.

\bibitem{pan2021mulls}
Y.~H. Z. S. Z.~L. Yue~Pan, Pengchuan~Xiao, ``Mulls: Versatile lidar slam via
  multi-metric linear least square,'' in \emph{IEEE International Conference on
  Robotics and Automation (ICRA)}.\hskip 1em plus 0.5em minus 0.4em\relax IEEE,
  2021.

\bibitem{yin2018locnet}
H.~Yin, L.~Tang, X.~Ding, Y.~Wang, and R.~Xiong, ``Locnet: Global localization
  in 3d point clouds for mobile vehicles,'' in \emph{2018 IEEE Intelligent
  Vehicles Symposium (IV)}.\hskip 1em plus 0.5em minus 0.4em\relax IEEE, 2018,
  pp. 728--733.

\bibitem{dube2020segmap}
R.~Dub{\'e}, A.~Cramariuc, D.~Dugas, H.~Sommer, M.~Dymczyk, J.~Nieto,
  R.~Siegwart, and C.~Cadena, ``Segmap: Segment-based mapping and localization
  using data-driven descriptors,'' \emph{The International Journal of Robotics
  Research}, vol.~39, no. 2-3, pp. 339--355, 2020.

\bibitem{Bai2020D3FeatJL}
X.~Bai, Z.~Luo, L.~Zhou, H.~Fu, L.~Quan, and C.~Tai, ``D3feat: Joint learning
  of dense detection and description of 3d local features,'' \emph{2020
  IEEE/CVF Conference on Computer Vision and Pattern Recognition (CVPR)}, pp.
  6358--6366, 2020.

\bibitem{muja2009fast}
M.~Muja and D.~G. Lowe, ``Fast approximate nearest neighbors with automatic
  algorithm configuration,'' in \emph{International Conference on Computer
  Vision Theory and Applications}, vol.~1.\hskip 1em plus 0.5em minus
  0.4em\relax Scitepress, 2009, pp. 331--340.

\bibitem{elsebergcomparison}
J.~Elseberg, S.~Magnenat, R.~Siegwart, and A.~N{\"u}chter, ``Comparison of
  nearest-neighbor-search strategies and implementations for efficient shape
  registration,'' \emph{Journal of Software Engineering for Robotics (JOSER)},
  vol.~3, no.~1, pp. 2--12, 2012.

\bibitem{kovalenko2019sensor}
D.~Kovalenko, M.~Korobkin, and A.~Minin, ``Sensor aware lidar odometry,'' in
  \emph{2019 European Conference on Mobile Robots (ECMR)}.\hskip 1em plus 0.5em
  minus 0.4em\relax IEEE, 2019, pp. 1--6.

\bibitem{YangPAMI2016}
J.~Yang, H.~Li, D.~Campbell, and Y.~Jia, ``Go-icp: A globally optimal solution
  to 3d icp point-set registration,'' \emph{IEEE Transactions on Pattern
  Analysis and Machine Intelligence}, vol.~38, no.~11, pp. 2241--2254, 2016.

\bibitem{GrangerECCV2002}
S.~Granger and X.~Pennec, ``Multi-scale em-icp: A fast and robust approach for
  surface registration,'' in \emph{European Conference on Computer Vision ---
  ECCV 2002}, A.~Heyden, G.~Sparr, M.~Nielsen, and P.~Johansen, Eds.\hskip 1em
  plus 0.5em minus 0.4em\relax Berlin, Heidelberg: Springer Berlin Heidelberg,
  2002, pp. 418--432.

\bibitem{YangTRO2021}
H.~Yang, J.~Shi, and L.~Carlone, ``Teaser: Fast and certifiable point cloud
  registration,'' \emph{IEEE Transactions on Robotics}, vol.~37, no.~2, pp.
  314--333, 2021.

\bibitem{6696380}
J.~{Saarinen}, H.~{Andreasson}, T.~{Stoyanov}, and A.~J. {Lilienthal}, ``Normal
  distributions transform monte-carlo localization (ndt-mcl),'' in \emph{2013
  IEEE/RSJ International Conference on Intelligent Robots and Systems}, 2013,
  pp. 382--389.

\bibitem{sun2020localising}
L.~Sun, D.~Adolfsson, M.~Magnusson, H.~Andreasson, I.~Posner, and T.~Duckett,
  ``Localising faster: Efficient and precise lidar-based robot localisation in
  large-scale environments,'' in \emph{2020 IEEE International Conference on
  Robotics and Automation (ICRA)}.\hskip 1em plus 0.5em minus 0.4em\relax IEEE,
  2020, pp. 4386--4392.

\bibitem{Oleynikova2016SignedDF}
H.~Oleynikova, A.~Millane, Z.~Taylor, E.~Galceran, J.~Nieto, and R.~Siegwart,
  ``Signed distance fields: A natural representation for both mapping and
  planning,'' in \emph{Robotics: Science and Systems 2016. Workshop: Geometry
  and Beyond - Representations, Physics, and Scene Understanding for Robotics}.

\bibitem{Perezgrau17JARS}
F.~Perez-Grau, F.~Caballero, A.~Viguria, and A.~Ollero, ``{Multi-sensor
  three-dimensional Monte Carlo localization for long-term aerial robot
  navigation},'' \emph{International Journal of Advanced Robotic Systems},
  vol.~14, no.~5, 2017.

\bibitem{Akai2020}
N.~{Akai}, T.~{Hirayama}, and H.~{Murase}, ``3d monte carlo localization with
  efficient distance field representation for automated driving in dynamic
  environments,'' in \emph{2020 IEEE Intelligent Vehicles Symposium (IV)},
  2020, pp. 1859--1866.

\bibitem{DuboisIros2020}
R.~Dubois, A.~Eudes, J.~Moras, and V.~Frémont, ``Dense decentralized
  multi-robot slam based on locally consistent tsdf submaps,'' in \emph{2020
  IEEE/RSJ International Conference on Intelligent Robots and Systems (IROS)},
  2020, pp. 4862--4869.

\bibitem{Pedrosa2017EfficientLB}
E.~Pedrosa, A.~Pereira, and N.~Lau, ``Efficient localization based on scan
  matching with a continuous likelihood field,'' in \emph{2017 IEEE
  International Conference on Autonomous Robot Systems and Competitions
  (ICARSC)}, 2017, pp. 61--66.

\bibitem{HornungIROS2010}
A.~Hornung, K.~M. Wurm, and M.~Bennewitz, ``Humanoid robot localization in
  complex indoor environments,'' in \emph{2010 IEEE/RSJ International
  Conference on Intelligent Robots and Systems}, 2010, pp. 1690--1695.

\bibitem{dantanarayana2016c}
L.~Dantanarayana, G.~Dissanayake, and R.~Ranasinge, ``C-log: A chamfer distance
  based algorithm for localisation in occupancy grid-maps,'' \emph{CAAI
  transactions on intelligence technology}, vol.~1, no.~3, pp. 272--284, 2016.

\bibitem{ceres-solver}
S.~Agarwal, K.~Mierle, and Others, ``Ceres solver,''
  \url{http://ceres-solver.org}.

\bibitem{airsim2017fsr}
\BIBentryALTinterwordspacing
S.~Shah, D.~Dey, C.~Lovett, and A.~Kapoor, ``Airsim: High-fidelity visual and
  physical simulation for autonomous vehicles,'' in \emph{Field and Service
  Robotics}, 2017. [Online]. Available: \url{https://arxiv.org/abs/1705.05065}
\BIBentrySTDinterwordspacing

\bibitem{Rusu_ICRA2011_PCL}
R.~B. Rusu and S.~Cousins, ``{3D is here: Point Cloud Library (PCL)},'' in
  \emph{{IEEE International Conference on Robotics and Automation (ICRA)}},
  Shanghai, China, May 9-13 2011.

\end{thebibliography}

\end{document}